\documentclass[letterpaper, 10 pt, conference]{ieeeconf}  

\IEEEoverridecommandlockouts                              
\title{How Shall I Drive? Interaction Modeling and Motion Planning towards Empathetic and Socially-Graceful Driving}

\author{Yi Ren$^{1}$, Steven Elliott$^{1}$, Yiwei Wang$^{1}$, Yezhou Yang$^{2}$, and Wenlong Zhang$^{3}$
\thanks{$^{1}$Yi Ren, Steven Elliott, and Yiwei Wang are with Department of Mechanical and Aerospace Engineering, Arizona State University, Tempe, AZ, 85287, USA. Email:
        {\tt\small yiren@asu.edu};
        {\tt\small stevenelliott10@gmail.com};
        {\tt\small yiwei.wang.3@asu.edu}}%
\thanks{$^{2}$Yezhou Yang is with the School of Computing, Informatics, and Decision Systems Engineering, Arizona State University, Tempe, AZ, USA. Email:
         {\tt \small  yz.yang@asu.edu}}%
\thanks{$^{3}$Wenlong Zhang is with The Polytechnic School, Ira A. Fulton Schools of Engineering, Arizona State University, Mesa, AZ, 85212, USA. Email: 
        {\tt\small wenlong.zhang@asu.edu}}%
}

\usepackage[utf8]{inputenc}
\usepackage{amssymb}
\usepackage{amsmath}
\usepackage{graphicx}  
\usepackage{xcolor}
\usepackage{hyperref}
\DeclareMathOperator*{\argmax}{arg\,max}
\DeclareMathOperator*{\argmin}{arg\,min}
\usepackage[linesnumbered,ruled]{algorithm2e} 
\usepackage{color}
\iftrue 
        \newcommand{\cutsectionup}{\vspace*{-0.1in}}
        \newcommand{\cutsectiondown}{\vspace*{-0.07in}}

        \newcommand{\cutsubsectionup}{\vspace*{-0.09in}}
        \newcommand{\cutsubsectiondown}{\vspace*{-0.06in}}

        \newcommand{\cutparagraphup}{\vspace*{-0.17in}}
        \newcommand{\cutparagraphdown}{\vspace*{-0.03in}}

        \newcommand{\cutcaptionup}{\vspace*{-0.1in}}
        \newcommand{\cutcaptiondown}{\vspace*{-0.2in}}

        \newcommand{\cuttablecaptiondown}{\vspace*{-0.15in}}

        \newcommand{\cutequationup}{\vspace*{-0.07in}}
        \newcommand{\cutequationdown}{\vspace*{-0.07in}}

        \newcommand{\cuttableup}{}
        \newcommand{\cuttabledown}{}

        \newcommand{\cut}{{\vspace*{-0.02in}}}
        \newcommand{\cutmore}{{\vspace*{-0.06in}}}
        \newcommand{\negcut}{}

\else 
        \newcommand{\cutsectionup}{}
        \newcommand{\cutsectiondown}{}

        \newcommand{\cutsubsectionup}{}
        \newcommand{\cutsubsectiondown}{}

        \newcommand{\cutparagraphup}{
        \newcommand{\cutparagraphdown}{}

        \newcommand{\cutcaptionup}{}
        \newcommand{\cutcaptiondown}{}

        \newcommand{\cutequationup}{}
        \newcommand{\cutequationdown}{}

        \newcommand{\cuttableup}{}
        \newcommand{\cuttabledown}{}

        \newcommand{\cut}{}
        \newcommand{\cutmore}{}
        \newcommand{\negcut}{}
\fi

\begin{document}

\maketitle
\thispagestyle{empty}
\pagestyle{empty}

\begin{abstract}
While intelligence of autonomous vehicles (AVs) has significantly advanced in recent years, accidents involving AVs suggest that these autonomous systems lack gracefulness in driving when interacting with human drivers. In the setting of a two-player game, we propose model predictive control based on \textit{social gracefulness}, which is measured by the discrepancy between the actions taken by the AV and those that could have been taken in favor of the human driver. We define \textit{social awareness} as the ability of an agent to infer such favorable actions based on knowledge about the other agent's intent, and further show that \textit{empathy}, i.e., the ability to understand others' intent by simultaneously inferring others' understanding of the agent's self intent, is critical to successful intent inference. Lastly, through an intersection case, we show that the proposed gracefulness objective allows an AV to learn more sophisticated behavior, such as passive-aggressive motions that gently force the other agent to yield.              
\end{abstract}

\section{Introduction}
\label{sec:intro}
\cutsectiondown
While intelligence of autonomous vehicles (AVs) has significantly advanced in recent years, accidents involving AVs suggested that these autonomous systems lack gracefulness in driving.
In one incident, Waymo reported that its AV (denoted by ``M'' hereafter) was stuck at an intersection because human drivers (denoted by ``H'') coming from the other road never stopped fully while approaching the intersection, making M believe that H wanted the right of way and causing it to stop indefinitely~\cite{waymocarstop}. The first claim against an AV manufacturer has been filed against General Motors in 2018, following a collision between a Cruise Automation 2016 Chevrolet Bolt and a motorcyclist, who alleged that the car swerved into his path without his agreement~\cite{baron_2018}.

Both incidents can be attributed to M's lack of understanding of H's intent and thus its failure to predict H's future motion. In interactions that involve both agents as decision makers, M's inference of H's intent depends on not only observations from H, but also knowledge about H's understanding of M. In other words, M needs to be empathetic in order to understand what leads to H's behavior. To elaborate, consider that an agent's intent is defined as parameters of its control objective. The interaction between M and H \textit{ from the perspective of M} can be modeled as results from a series of games along time, with each new game parameterized by M's intent, and M's current inference of H's intent. M chooses motion based on the perceived game and its \textit{planning strategy} (e.g., reactive or proactive planning, see Sec.~\ref{sec:planning} for definitions), and assumes that H follows the same decision making process for motion planning, with its own perception of the game and planning strategy. Under this game-theoretic setting, the inference of H's intent is in conjunction with that of H's inference of M's intent, i.e., what H expects M to do, which is in general different from M's true intent. Enabling M's awareness of the difference between M's and H's perception of the interaction is what makes M empathetic. 

\begin{figure}
    \centering
    \includegraphics[width=0.8\linewidth]{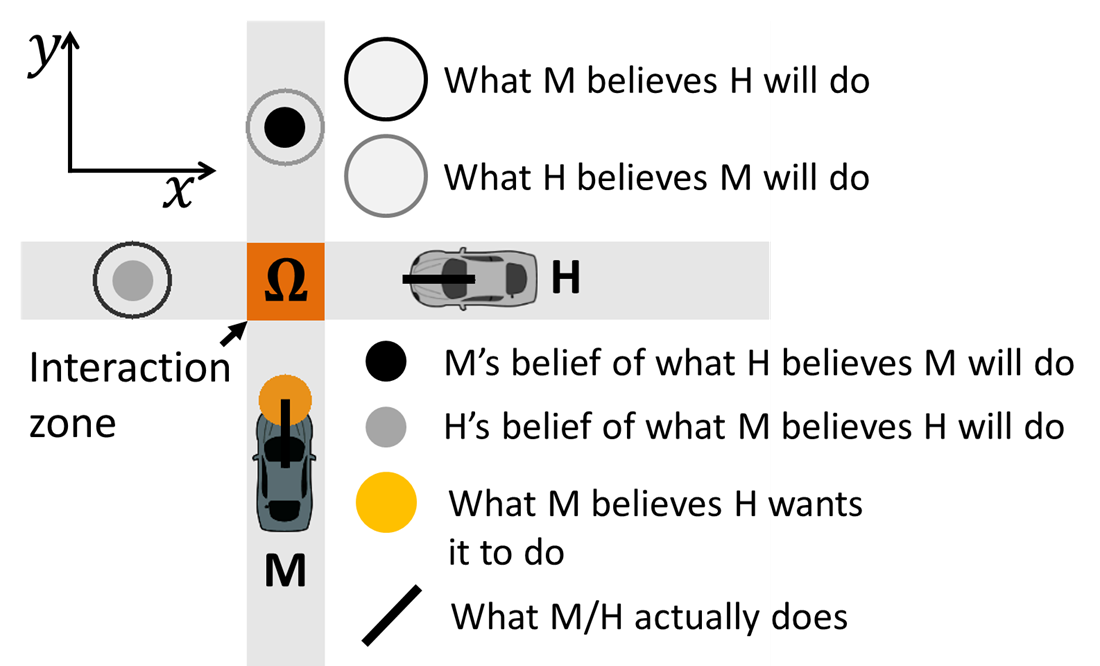}
    \vspace{-10pt}
    \caption{An illustration of social awareness of an agent. By inferring H's intent, M infers how H will move, how M is expected to move, and what movements of M will be in favor of H.}
    \label{fig:example}
\vspace{-20pt}
\end{figure}

Building upon this, this paper further introduces a motion planning strategy that incorporates \textit{social gracefulness}, which is defined as the discrepancy between the actions taken by M and those that could have been taken in favor of H. Take the intersection case in Fig.~\ref{fig:example} for example. When H moves slowly, an empathetic M knows that he is expected to take the right of way. A socially-aware M, in addition, knows that himself driving slowly to allow H to pass would do H a favor. Note that to acquire social awareness, M would need to understand H's intent, and thus be empathetic in the first place. We will show that considering the gracefulness objective in planning allows M to learn more sophisticated behavior, such as passive aggressive motions that gently force H to yield.         

\begin{figure*}[ht!]
    \centering
    \includegraphics[width=0.8\textwidth]{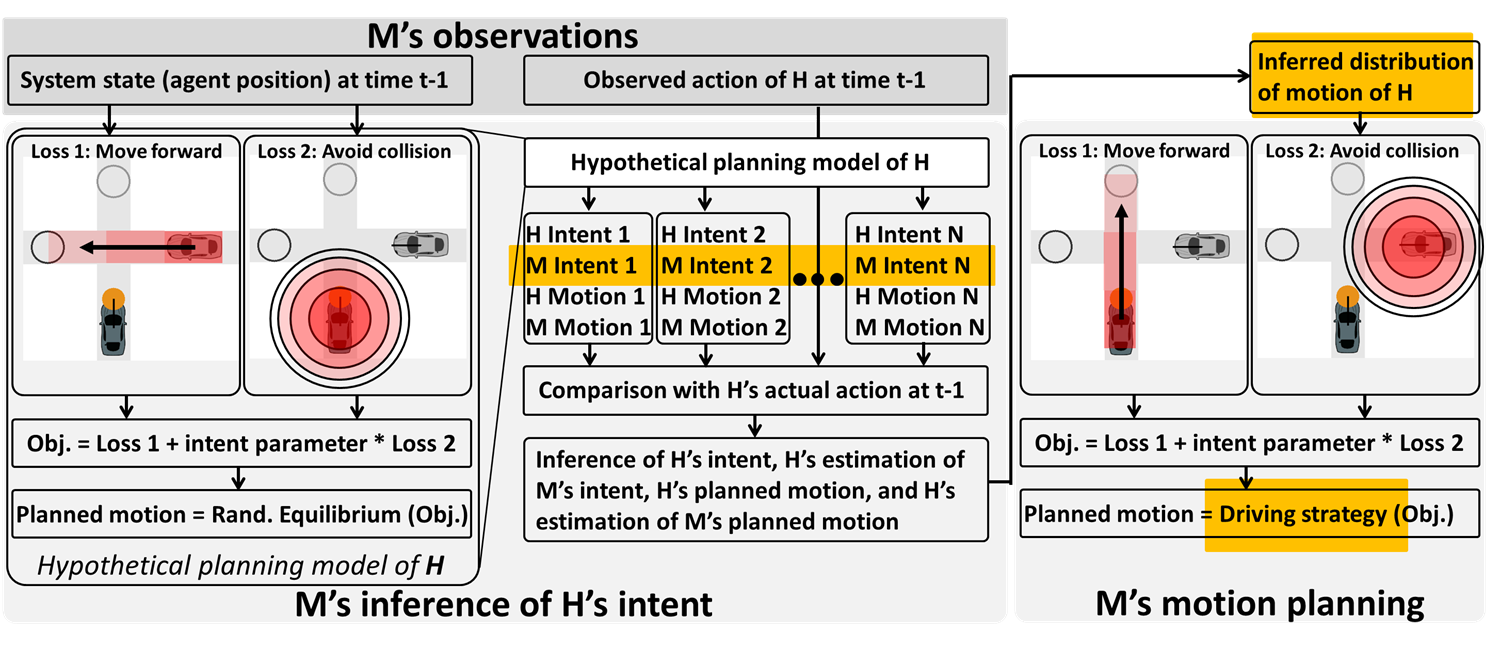}
    \vspace{-20pt}
    \caption{Summary of the intent inference and motion planning algorithms of M. To perform intent inference, M needs to model H's motion planning hypothetically. The highlighted modeling elements include (1) the game-theoretic intent inference algorithm that enables empathy, (2) the incorporation of inference uncertainty in motion planning, and (3) more sophisticated motion planning objectives enabled by intent inference.}
    \label{fig:summary}
\vspace{-20pt}
\end{figure*}
The technical approaches of the paper can be summarized in two steps: In Sec.~\ref{sec:intent}, we formulate the intent inference problem where the intents of both agents are jointly inferred, based on the assumption that the both agents model each other to adopt a baseline planning strategy where the agent plans motions based on the set of Nash equilibria of the game perceived by the others. 
In Sec.~\ref{sec:planning}, we explore more sophisticated planning strategies enabled by the inferred intents, and evaluate how these strategies influence the actual social gracefulness and social efficiency of an interaction between H and M. 
Three driving strategies will be explored: (1) A \textit{reactive} M plans based on the inferred distribution of future motions of H. (2) A \textit{proactive} M exploits the fact that H's future motion are dependent on M's. 
(3) A \textit{socially-aware} M is built upon the proactive model, with the additional gracefulness loss.
The proposed intent inference and motion planning models are summarized in Fig.~\ref{fig:summary}.

The intellectual contributions of this paper are two-fold: First, we propose mechanistic intent inference and motion planning algorithms for two-agent interactions. Our approach is different from existing studies where the control policy or objective of either M or H is learned offline through data. The introduction of empathy was previously discussed in Liu et al.~\cite{liu_who_2016}. This idea is extended to non-collaborative scenarios with non-quadratic control objectives. Second, we introduce the definition of social gracefulness as a performance measure for human-machine interactions. The idea of measuring risk through the difference between the driver's actions and what he is expected to do according to traffic rules was introduced in \cite{lefevre2013intention}. We take a parallel approach by defining the nominal actions as shaped by others' intents (i.e., tacit social norms), rather than by explicit traffic rules. 
\section{Related Work}
\cutsectiondown
\label{sec:lit}
AV has been an active and fast developing research topic in the past few years. Scenarios such as lane changing, lane merging, and traffic intersections, where AVs closely interact with human-driving vehicles, are widely investigated~\cite{liu_distributed_2018, zhang_finite_2017,lefevre_survey_2014}. Human drivers are often modeled as rational and intelligent decision makers in these studies. Pentland and Liu proposed a dynamic Markov models to recognize and predict human driving actions~\cite{pentland1999modeling}. Levine et al. adopted an inverse reinforcement learning method to explore human drivers' reward functions based on human demonstration recorded through a driving simulator~\cite{levine2010feature}. Machine learning is used to predict the lateral motions and braking actions of human driver~\cite{wang_learning_2018,okamoto2018comparative,tang2018lane}. These data-driven approaches require a large amount of data for model training. Wang et al. discussed the appropriate amount of data to achieve sufficient understanding of human driving behavior~\cite{wang2017much}.

Based on proper human driver models, an essential question remains as how to achieve safe and efficient human-machine interactions. Nikolaidis et al.~\cite{nikolaidis_human-robot_2017} proposed a mutual adaptation algorithm where human's willingness to adapt is inferred and incorporated into the robot's motion planning. 
In a similar two-agent collaborative setting, Liu et al.~\cite{liu_who_2016} proposed a ``Blame-all strategy'' for intent inference. Much like this paper, ``Blame-all'' performs inference of others' intents by taking into account others' inference of the agent's intent, with the limited scope of quadratic control objectives. 
For dynamically changing agent parameters (such as intents), Foerster et al.~\cite{foerster_learning_2017} proposed Learning with Opponent-Learning Awareness (LOLA), a method for agents to incorporate the anticipated learning of other agents during its own adaptation (similar to \cite{sadigh_planning_2016} and the proactive vs. reactive case of this paper). LOLA, however, had limited discussion on scenarios where opponent parameters and parameter tuning schemes need to be inferred in a higher level game (e.g., proactive vs. proactive as in this paper). Specific to AVs, various game theoretic models have been constructed to describe different interactions between human drivers and AVs~\cite{yu_human-like_2018,carlino_auction-based_2013}. 
In a setting of partially observable Markov decision process, Li et al. introduced a ``Level-k'' reasoning model to evolve the strategies that agents adopt in driving interactions~\cite{li2018game}. Nonetheless, their solution requires an iterative search through reinforcement learning. In contrast, mechanistic models (e.g., \cite{liu_who_2016,nikolaidis_human-robot_2017} and this paper) are one-shot and are more suitable for scenarios where agents are required to learn and adapt in real time. 
\section{Intent Inference}
\label{sec:intent}
\cutsectiondown
This section introduces the motion planning game and the inverse problem of intent inference. For simplicity, we focus on an interaction between two agents, M and H. The extension to multi-agent settings will be left as future work. 
The inference by M is based on the assumptions that H uses a \textit{baseline} motion planning strategy (see Sec.~\ref{sec:baseline}), and that H believes that M uses the same strategy. 

\cutsubsectionup
\subsection{The motion planning game}
\label{sec:game}
\cutsubsectiondown
Let $\textbf{s}_i(t)$ and $\textbf{u}_i(t)$ be the \textit{state} and \textit{action} at time $t$, with the subscript $i \in \{H,M\}$ specifying the agent. We consider deterministic state transition: $\textbf{s}(t) = \textbf{s}(t-1) + \textbf{u}(t-1)$, and define $i$'s planned \textit{motion} at $t$ as a sequence of actions $\xi_i(t) := [\textbf{u}_i(t), \textbf{u}_i(t+1), \cdots, \textbf{u}_i(t+T-1)]$, where $T$ is a predefined finite time horizon. The dependency on $t$ will be omitted for brevity.
The planning game is formulated as follows: The instantaneous loss of agent $i$ is 
\cutequationup
\begin{equation}
    l(\xi_i,\xi_j,c_i,\textbf{s}_i,\textbf{s}_j) = l_{\text{safety}}(\xi_i,\xi_j,\textbf{s}_i,\textbf{s}_j) + c_i l_{\text{task}}(\xi_i,\textbf{s}_i),
\cutequationdown
\end{equation}
where $l_{\text{safety}}$ measures the safety loss, $l_{\text{task}}$ the fulfillment of a task, e.g., ``lane changing'' or ``crossing the intersection'' (see Sec.~\ref{sec:case}), and the intent parameter $c_i$ governs the behavior of the vehicle. Note that $l_{\text{safety}}$ involves the distance between $i$ and $j$, and thus requires $\xi_i$, $\xi_j$, $\textbf{s}_i$, and $\textbf{s}_j$ as inputs. The dependency on $\textbf{s}_i$ and $\textbf{s}_j$ will be omitted for brevity.
The payoff for $i$ in the game is defined as the accumulative loss 
\cutequationup
\begin{equation}
    f(\xi_i,\xi_j,c_i) = \sum_{\tau=t}^{t+T-1}l(\xi_i,\xi_j,c_i)
\cutequationdown
\end{equation}
for $\xi_i$ that satisfies $\textbf{g}_i(\xi_i)\leq \textbf{0}$, which represents boundaries on future states. For infeasible $\xi_i$, $f=\infty$. 
Each agent chooses its motion from a finite candidate set $\Xi_i$. Under this setting, the game is guaranteed to have at least one Nash equilibrium (Theorem 23 in \cite{jiang2009tutorial}). 

Denote $\hat{c}_j$ as $i$'s estimation of $j$'s intent (see Sec.~\ref{sec:inference} for the inference of $\hat{c}_j$). The game at time $t$ yields a Nash equilibrium set $\mathcal{Q}(c_i,\hat{c}_j,t)=\{(\xi_i^*, \hat{\xi}_j^*)\}$, where $\hat{\xi}_j^*$ is $i$'s estimation of $j$'s planned motion. Each element of $\mathcal{Q}$ satisfies
\cutequationup
\begin{equation}
    \begin{aligned}
        \xi_i^* = \underset{\{\xi_i \in \Xi_i; \textbf{g}_i(\xi_i)\leq \textbf{0}\}}{\arg\min} f(\xi_i,\hat{\xi}_j^*,c_i),\\
        \hat{\xi}_j^* = \underset{\{\xi_j \in \Xi_j; \textbf{g}_j(\xi_j)\leq \textbf{0}\}}{\arg\min} f(\xi_j,\xi_i^*,\hat{c}_j).
    \end{aligned}
    \label{eq:mpc}
\cutequationdown
\end{equation}
For example, at an intersection without a stop sign, either M or H yielding to the other will satisfy Eq.~\eqref{eq:mpc} when both M and H are non-aggressive (i.e., with small $c$). It is important to note that for $i$ to understand $j$, $i$ needs to put himself in the shoes of $j$ and derive the equilibrium set from $j$'s perspective. Since $j$ does not know $c_i$, a necessary correction is to 
replace $c_i$ with $\tilde{c}_i$, i.e., $i$'s estimation of $j$'s estimation of $i$'s intent, and similarly, $\xi_i^*$ with $\tilde{\xi}_i^*$.

\cutsubsectionup
\subsection{The baseline driving strategy}
\label{sec:baseline}
\cutsubsectiondown
Agent $i$ is modeled to believe that $j$ plans its motion at $t$ by sampling \textit{uniformly} from $\mathcal{Q}(\tilde{c}_i,\hat{c}_j,t)$, i.e., $j$'s motion follows the probability mass function:
\cutequationup
\begin{equation*}
    p(\hat{\xi}_j;\tilde{c}_i,\hat{c}_j,t) \propto |\{\hat{\xi}_j; (\tilde{\xi}_i, \hat{\xi}_j) \in \mathcal{Q}(\tilde{c}_i,\hat{c}_j,t) \}|,
\cutequationdown
\end{equation*}
where $|\cdot|$ is the cardinality of a set.

\cutsubsectionup
\subsection{Inference of intent and motion}
\label{sec:inference}
\cutsubsectiondown
Eq.~\eqref{eq:intent} formulates the inference problem where we estimate the intents ($\tilde{c}_i$ and $\hat{c}_j$) and the planned motions ($\tilde{\xi}_i^*$ and $\hat{\xi}_j^*$) at time $t$ based on the states and actions of $i$ and $j$ at $t-1$. The idea is that the estimated intents should be such that $j$'s planned motion at $t-1$ with the highest probability mass (denoted by $\hat{\xi}_j^{\dagger}(t-1)$) should have its first action (denoted by $\hat{\textbf{u}}_j^{\dagger}(t-1)$) matching the observed action of $j$ at $t-1$ ($\textbf{u}_j(t-1)$), or verbally, what $j$ should have done should match with what it actually did.
\cutequationup
\begin{equation}
    \begin{aligned}
        \min_{\tilde{c}_i, \hat{c}_j} & \quad ||\hat{\textbf{u}}_j^{\dagger}(t-1) - \textbf{u}_j(t-1)||_2^2 \\
        \text{subject to} & \quad \hat{\xi}_j^{\dagger}(t-1) = \argmax_{\xi_j \in \Xi_j} p(\xi_j;\tilde{c}_i,\hat{c}_j,t-1)
    \end{aligned}
    \label{eq:intent}
\cutequationdown
\end{equation}
We use a finite intent set $\mathcal{C}$ for both agents to represent different levels of aggressiveness in driving. Since both $\mathcal{C}$ and $\Xi_{i,j}$ are finite, the inference is done through an enumeration over the solution space $\mathcal{C} \times \mathcal{C}$. The outcome of the enumeration is a set $\mathcal{S}(t) = \left\{(\tilde{c}_i^*, \hat{c}_j^*)_k\right\}_{k=1}^{K}$, where each element is a global solution to Eq.~\eqref{eq:intent}. To quantify the uncertainty in inference and later incorporate it into motion planning, we assign equal probability mass ($1/K$) to each solution, based on which we can compute the empirical joint distribution $p(\tilde{c}_i, \hat{c}_j; t)$ and the marginals $p(\tilde{c}_i;t)$ and $p(\hat{c}_j;t)$, by counting the appearances of all elements of $\mathcal{C}$ in $\mathcal{S}(t)$. 
From the joint distribution $p(\tilde{c}_i, \hat{c}_j; t)$, we can infer $j$'s planned motion. From each $(\tilde{c}_i^*, \hat{c}_j^*) \in \mathcal{S}(t)$ we can compute a conditional distribution of $j$'s motion starting from $t-1$, denoted by $p(\hat{\xi}_j;\tilde{c}^*,\hat{c}_j^*,t-1)$, through $\mathcal{Q}(\tilde{c}_i^*,\hat{c}_j^*,t-1)$. We can then calculate the marginal $p(\hat{\xi}_j;t-1)$ using $p(\hat{\xi}_j;\tilde{c}^*,\hat{c}_j^*,t-1)$ and $p(\tilde{c}_i, \hat{c}_j; t-1)$. Note that this is the distribution of $j$'s motion at $t-1$ rather than $t$ since we formulated the game at $t-1$. Although one can formulate a new game at $t$ 
to derive $p(\xi_j;t)$, in this paper we will approximate $p(\xi_j;t)$ using $p(\xi_j;t-1)$. Similarly, we can also infer $i$'s planned motion as expected by $j$, denoted by $p(\tilde{\xi}_i;t)$. We shall emphasize that $p(\tilde{\xi}_i;t)$ is not the distribution of motion that $i$ will follow, rather, it represents $i$'s awareness of what $j$ expects it to do.

\cutsubsectionup
\subsection{Leveraging past observations}
\cutsubsectiondown
Note that if we consider $c_j$ to be time invariant during the interaction, 
then all previous observations can be leveraged to infer $c_j$, leading to a modified problem:
\cutequationup
\begin{equation}
    \begin{aligned}
        \min_{\{\tilde{c}_i(\tau)\}_{\tau=1}^{t-1}, \hat{c}_j} & \quad \sum_{\tau=1}^{t-1} ||\hat{\textbf{u}}_j^{\dagger}(\tau) - \textbf{u}_j(\tau)||_2^2 \\
        \text{subject to} & \quad \xi_j^{\dagger}(\tau) = \argmax_{\xi_j \in \Xi_j} p(\xi_j;\tilde{c}_i(\tau),\hat{c}_j,\tau) 
        ~\forall \tau.
    \end{aligned}
    \label{eq:intent2}
\cutequationdown
\end{equation}
We allow $\tilde{c}_i$ to freely change along time since $j$ may change its mind about $i$.
Solutions to Eq.~\eqref{eq:intent2}, denoted by $\bar{\mathcal{S}}(t)$, can be found in a recursive way based on solutions to Eq.~\eqref{eq:intent}. Consider an intent candidate $\hat{c}_j$ that exists in both $\bar{\mathcal{S}}(t-1)$ and $\mathcal{S}(t)$. Let solutions in $\bar{\mathcal{S}}(t-1)$ that contain $\hat{c}_j$ be in the form of $(\textbf{a}, \hat{c}_j)$, where $\textbf{a} = [a_1,\cdots,a_{t-1}] \in \mathcal{A} \subset \mathcal{C}^{t-1}$ is an array of intents of $i$. Similarly, let solutions in $\mathcal{S}(t)$ that contain $\hat{c}_j$ be in the form of $(b, \hat{c}_j)$, where $b \in \mathcal{B}$ is an intent of $i$. Also denote the operation of appending element $b$ to the array $\textbf{a}$ as $[\textbf{a},b]$. Then $([\textbf{a},b], \hat{c}_j)$ for all $\textbf{a} \in \mathcal{A}$ and $b \in \mathcal{B}$ is a solution to Eq.~\eqref{eq:intent2} at $t$. Following this property, we have $\bar{p}(\hat{c}_j;t) \propto \bar{p}(\hat{c}_j;t-1)p(\hat{c}_j;t)$. 
The update of $\bar{p}(\hat{c}_j;t)$ will trigger that of the joint probability $p(\tilde{c}_i, \hat{c}_j; t)$ and the marginal $p(\hat{\xi}_j;t)$. Their updated counterparts are denoted by $\bar{p}(\tilde{c}_i, \hat{c}_j; t)$ and $\bar{p}(\hat{\xi}_j;t)$, respectively.
\cutsectionup
\section{Motion Planning}
\label{sec:planning}
\cutsectiondown
We introduce three planning strategies that incorporate the inferred intents and motions. 
Note that in this paper, agents are only modeled to infer others' intents rather than their planning strategies, i.e., they are assumed to use the baseline strategy but may in fact use different ones (see Fig.~\ref{fig:summary}). Potential formulations of higher-level inference algorithms will be discussed in Sec.~\ref{sec:discussion}.
\cutsubsectionup
\subsection{Reactive motion} 
\cutsubsectiondown
Given the distribution of $j$'s future motions $\bar{p}(\hat{\xi}_j;t)$, a reactive agent plans its motion by minimizing the expected loss within a time window:
\cutequationup
\begin{equation*}
    \begin{aligned}
        \min_{\xi_i \in \Xi_i, \textbf{g}_i(\xi_i) \leq \textbf{0}} & \quad \mathbb{E}_{\bar{p}(\hat{\xi}_j;t)} f(\xi_i,\hat{\xi}_j,c_i)
    \end{aligned}
    \label{eq:reactive}
\cutequationdown
\end{equation*}

\cutsubsectionup
\subsection{Proactive motion}
\cutsubsectiondown
A proactive agent $i$ plans by taking into consideration the dependency of $j$'s planning on $i$'s next action. Specifically, $i$ calculates the conditional distribution $\bar{p}(\hat{\xi}_j;\xi_i,t)$ based on $\xi_i$ and $\bar{p}(\hat{c}_j; t)$, assuming that $j$ will quickly respond to $\xi_i$. To do so, $i$ first finds the set of optimal motions of $j$ for every $\hat{c}_j \in \mathcal{C}$ given $\xi_i$. This set is denoted by 
$\mathcal{Q}_j(\xi_i)= \cup_{\hat{c}_j \in \mathcal{C}} \mathcal{Q}_j(\xi_i,\hat{c}_j)$ where $\mathcal{Q}_j(\xi_i,\hat{c}_j)= \{\hat{\xi}_j^*; \hat{\xi}_j^* = \argmin_{\xi_j \in \Xi_j} f(\xi_j, \xi_i, \hat{c}_j)\}$.
Then for each element $\hat{\xi}_j^* \in \mathcal{Q}_j(\xi_i)$, we can compute 
\cutequationup
\begin{equation*}
    \bar{p}(\hat{\xi}_j^*;\xi_i,t) = \sum_{\hat{c}_j \in \mathcal{C}} \frac{\bar{p}(\hat{c}_j;t)1(\hat{\xi}_j^* \in \mathcal{Q}_j(\xi_i,\hat{c}_j))}{|\mathcal{Q}_j(\xi_i,\hat{c}_j)|},
    \label{eq:pro3}
\cutequationdown
\end{equation*}
where $1(\cdot)$ is an indicator function. For $\hat{\xi}_j \in \Xi_j / \mathcal{Q}_j(\xi_i)$, we set $\bar{p}(\hat{\xi}_j;\xi_i,t)=0$.
We can now develop the planning problem for a proactive agent:
\cutequationup
\begin{equation*}
    \begin{aligned}
        \min_{\xi_i \in \Xi_i, \textbf{g}_i(\xi_i) \leq \textbf{0}} & \quad \mathbb{E}_{\bar{p}(\hat{\xi}_j;\xi_i,t)} f(\xi_i,\hat{\xi}_j,c_i)
    \end{aligned}
    \label{eq:proactive}
\cutequationdown
\end{equation*}

\cutsubsectionup
\subsection{Socially-aware motion} 
\cutsubsectiondown
To develop the planning problem for a socially-aware agent $i$, we will first find the set of motions $\{\xi_i^j\} \subset \Xi_i$ that are wanted by $j$ at time $t$. By definition, $\xi_i^j$ belongs to motion tuples $(\xi_i, \xi_j) \in \mathcal{Q}(\tilde{c}_i, \hat{c}_j, t)$ that minimizes $f(\xi_j, \xi_i, \hat{c}_j)$. This set is denoted by $\mathcal{Q}^j(\tilde{c}_i, \hat{c}_j, t)$. 
Based on $\mathcal{Q}^j(\tilde{c}_i, \hat{c}_j, t)$ and $\bar{p}(\tilde{c}_i, \hat{c}_j; t)$, we can calculate the conditional probability $p(\xi_i^j; \tilde{c}_i, \hat{c}_j, t)$ and the marginal $\bar{p}(\xi_i^j; t)$.
A socially-aware agent solves the following problem where the difference between $i$'s planned motion and the wanted motion from $j$ is added to the proactive loss:
\cutequationup
\begin{equation*}
    \begin{aligned}
        \min_{\xi_i \in \Xi_i, \textbf{g}_i(\xi_i) \leq \textbf{0}} & \quad  \mathbb{E}_{p(\hat{\xi}_j;\xi_i,t)} \left( f(\xi_i,\hat{\xi}_j,c_i) + \beta \mathbb{E}_{\bar{p}(\xi^j_i; t)} ||\xi^j_i - \xi_i||_2^2  \right)
    \end{aligned}
    \label{eq:social}
\cutequationdown
\end{equation*}
The weight $\beta \geq 0$ tunes the importance of social awareness in the objective: A large $\beta$ will make $i$ follow what it believes that $j$ wants it to do, while a small $\beta$ will convert $i$ to a proactive agent. 


\section{Simulation Cases and Analyses}
\label{sec:case}
In this section we will empirically answer two questions: (1) How do planning strategies influence interaction dynamics, 
and (2) why empathy is important for intent inference? 
We base all experiments on a simulated intersection case introduced as follows.

\cutsubsectionup
\subsection{Simulation setup}
\cutequationdown
The intersection case is summarized in Fig.~\ref{fig:example}, where M moves up (along y-axis) and H moves left (along x-axis). {\bf Motion representation}: With minor abuse of notation, we define $\xi_i$ as a scalar that determines the distance covered by an agent in $T=100$ steps, and assume that each agent moves at a uniform speed and a predefined direction: $\textbf{u}_M = (0.0, \xi_M/T)$ and $\textbf{u}_H = (-\xi_H/T, 0.0)$. The candidate motion set is $\Xi_i = \Xi_j = \{-1.0, 0.0, 1.0, 2.0, 3.0, 4.0, 5.0\}$, where $-1.0$ means backing up. {\bf Initial conditions}: M and H are initially located at $\textbf{s}_M(0) = (0.0,-2.0)$ and $\textbf{s}_H(0) = (2.0,0.0)$. The initial motions are set to $\xi_M(0) = \xi_H(0) = 5.0$. {\bf Losses}: $l_{\text{task}}$ penalizes the agent if it fails to move across the intersection within $T$ steps. Taking $M$ as an example, the loss is defined as $l_{\text{M,task}} = \exp(-s_M^{(x)}(t+T-1)+0.4)$, where $s_M^{(x)}$ is the state of M in the $x$ direction. 
The safety loss is defined as $l_{\text{safety}} = \exp(a(-D^2+b))$, where $D = ||\textbf{s}_M - \textbf{s}_H||_2^2$ when both cars are in the interaction area $\Omega$ as shown in Fig.~\ref{fig:example}, and otherwise $D = \infty$; $a=5.0$ so that the safety penalty increases significantly as the two cars approach each other; $b=1.5w^2$ where $w=1.33$ is the length of the car. This setting creates a safe zone for an agent since the penalty diminishes when $D^2 > b$. For the socially-graceful agent, $\beta$ is set to 0.1 for the following experiments. A parametric study on $\beta$ will be discussed later in this section. 
{\bf Performance metrics}: We measure the performance of a pair of motion planning strategies, denoted by $P_i \in \{\text{reactive},\text{proactive},\text{social}\}$ for $i \in \{M, H\}$, using two metrics, namely, social gracefulness and social efficiency. Gracefulness, denoted by $q_{\text{grace}}$, is defined as
\cutequationup
\begin{equation*}
q_{\text{grace}} = \sum_{t = 0}^{T_s} \mathbb{E}_{p(\xi_M^{\star};t)}|| \textbf{u}_M^{P_M}(t) - \textbf{u}_M^{\star} ||_2^2,
\cutequationdown
\end{equation*}
where $T_s$ is the time horizon for the entire simulation, $\textbf{u}_M^{P_M}(t)$ is the action taken by M at $t$ using strategy $P_M$, and $\textbf{u}_M^{\star}$ is the action that H actually wants M to perform, which can be different from what M believes that H wants it to perform. 
We only measure gracefulness of M (the AV). Social efficiency, denoted by $q_{\text{eff}}$, is defined as the time required for the agents to reach an agreement. An agreement is reached at time $t$ when what $j$ expects $i$ to do matches with what $i$ actually does, i.e., $p(\tilde{\xi}_i=\xi_i(t);t) =1$. Therefore we can define
\cutequationup
\begin{equation*}
    q_{\text{eff}} = \argmin_{t=\{0,\cdots,T_s\}, p(\tilde{\xi}_i=\xi_i(t);t)=1, i \in\{M,H\}} t.
\cutequationdown
\end{equation*}
{\bf Experiment setup}: The experiments will enumerate over $P_i \times P_j \in \{\text{reactive},\text{proactive},\text{social}\}^2$ and $c_M \times c_H \in \{1, 10^9\}^2$. $c_i = 1$ represents a non-aggressive agent that will try to avoid being in $\Omega$ with the other agent; $c_i = 10^9$ an extremely aggressive agent that is willing to collide with the other agent. To test whether an agent can handle unexpected intents, the candidate intent set for decoding is set to $\mathcal{C} = \{1, 10^3\}$. $c_i=10^3$ represents an aggressive agent that avoids collision. 
{\bf Conflict in inference}: It is worth mentioning that since the agents do not follow the baseline algorithm for choosing a motion, their motions do not necessarily belong to an equilibrium set. This causes occasions where the probability mass of the true intent of $j$ becomes zero as inferred by $i$, or where the updated probability distribution, e.g., $\bar{p}(\hat{c}_j;t)$, has all zero entries when all candidate intents have been eliminated during the inference process. In the implementation, we set $\bar{p}(\hat{c}_j;t)$ back to a uniform distribution when the latter happens. 

\cutsubsectionup
\subsection{Influence of planning strategy on the interaction}
\begin{figure}
    \centering
    \includegraphics[width=1.0\linewidth]{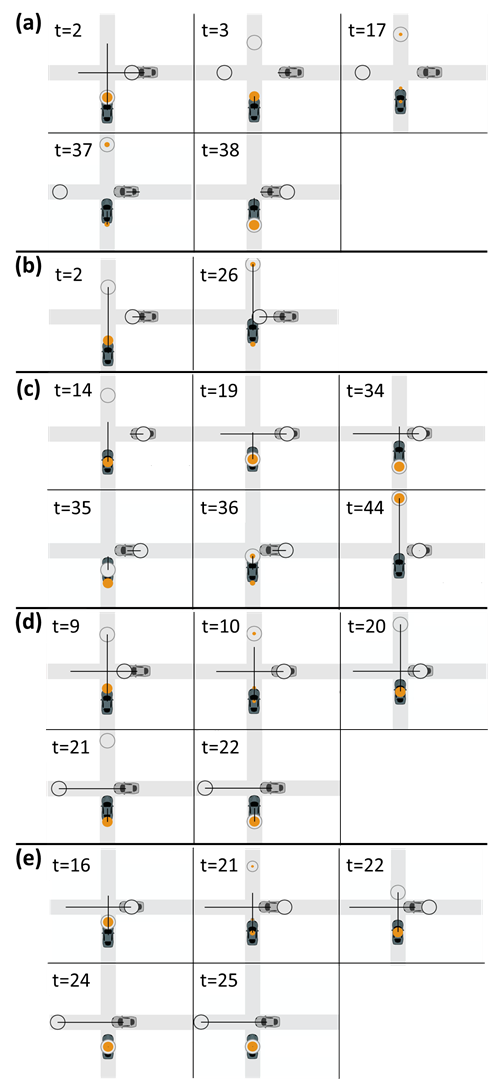}
    \caption{Snapshots of the interactions. (a) Reactive M vs. reactive H, $c_M = c_H = 1$. Both agents take the same actions at every time step since they share the same settings. After the initial action at $t=0$, both take the smallest step $\xi = 1$ at $t=1$ due to the concern that the other agent will take $\xi=5$. This iteration between two action values continues indefinitely. This deadlock can be broken when the agents have significantly different $c$s. (b)
    Proactive M vs. reactive H, $c_M = c_H = 1$.
    M successfully tricks H into believing that M is aggressive, and forces H to yield. (c) Graceful M vs. reactive H, $c_M = c_H = 1$. M reaches the interaction zone slightly before H, gently forcing H to yield. (d) Proactive M vs. reactive H, $c_M = 1$, $c_H = 10^9$. M reaches the interaction zone and realizes that it has to back off. (e) Graceful M vs. reactive H, $c_M = 1$, $c_H = 10^9$. M reaches the interaction zone after H due to its tendency to improve gracefulness, and therefore avoids sudden change in action. Legend follows Fig.~\ref{fig:example}.}
    \label{fig:result_summary}
\end{figure}

\cutsubsectiondown
We use Table~\ref{tab:exp1} to summarize gracefulness and efficiency under all combinations of planning strategies and intent settings. Due to page limit, we highlight two key findings. (1) The \textbf{socially-aware strategy balances efficiency and gracefulness}, provided that human drivers are reactive~\cite{costa2009comparing} and non-aggressive ($c_H=1$). In particular, a socially-aware M takes advantage of a non-aggressive agent, and thus it makes the interaction more efficient than a reactive agent (which leads to a stagnation as seen in Fig.~\ref{fig:result_summary}a), while at the same time in a more graceful way than a proactive agent (comparison between Fig.~\ref{fig:result_summary}b and c).
(2) In comparison with proactive planning, the \textbf{socially-aware M avoids ``panic reactions'' when facing unexpectedly aggressive H} during an interaction. As seen in Fig.~\ref{fig:result_summary}d, a proactive M realizes that it needs to back off \textit{after} it enters the interaction zone, forcing itself to make large change in its motion. In comparison, a socially-aware M avoids this as it has been partially complying to H's intent all along the way, thus creating a smoother negotiation with H, see Fig.~\ref{fig:result_summary}e. 

\begin{table}
\centering
\caption{$q_{\text{grace}}$ and $q_{\text{eff}}$ ($c_H=1$, $c_H=10^9$), lower values are better)}
\cuttablecaptiondown
\begin{tabular}{|c|c|c|c|}
    \hline
    \multicolumn{4}{|l|}{$q_{\text{grace}}\times 10^{-3}$}\\ 
    \hline
    & reactive H & proactive H & social H\\
    \hline
    reactive M & ($\infty$, $0.0$) & ($0.0$, $0.0$) & ($35.5, 0.0$) \\
    \hline
    proactive M & ($58.3$, $35.1$) & ($\infty$, $35.1$) & ($50.3$, $35.1$) \\
    \hline
    social M & ($31.2$, $12.5$) & ($12.5$, $12.5$) & ($\infty$, $12.5$)\\
    \hline
    \multicolumn{4}{|l|}{$q_{\text{eff}}$} \\
    \hline
    reactive M & ($\infty$, $1$) & ($1$, $1$) & ($44$, $1$) \\
    \hline
    proactive M & ($1$, $22$) & ($\infty$, $22$) & ($24$, $22$) \\
    \hline
    social M & ($44$, $24$) & ($24$, $24$) & ($\infty$, $24$)\\
    \hline
\end{tabular}
\label{tab:exp1} 
\vspace{-10pt}
\end{table}

\cutsubsectionup
\subsection{Parametric study on $\beta$}
\cutsubsectiondown
We performed a parametric study on the gracefulness weight $\beta$ to examine its influence on the gracefulness and efficiency of the interaction. Results are summarized in Table~\ref{tab:par}. In all cases, H is reactive and non-aggressive. As $\beta$ increases, M improves its gracefulness by taking actions closer to what is wanted by H. The efficiency, however, becomes worse towards $\beta = 0.5$ as the interaction becomes similar to that of reaction vs. reaction. Verbally, being graceful but not enough graceful will cause confusion in interactions. As $\beta$ further increases, M converges to follow a motion that is in favor of H, and thus improves the efficiency.  

\begin{table}
\centering
\caption{Parametric study on the gracefulness weight $\beta$}
\cuttablecaptiondown
\begin{tabular}{|c|c|c|c|c|c|c|}
    \hline
    $\beta$ & $0.05$ & $0.10$ & $0.15$ & $0.30$ & $0.50$ & $0.70$\\
    \hline
    $q_{\text{grace}}\times10^{-3}$ & $39.4$ & $31.2$ & $14.8$ & $11.2$ & $5.0$ & $2.2$ \\
    \hline
    $q_{\text{eff}}$ & $43$ & $44$ & $45$ & $49$ & $57$ & $20$ \\
    \hline
    right-of-way & M & M & H & H & H & H \\
    \hline
\end{tabular}
\label{tab:par}
\vspace{-20pt}
\end{table}

\cutsubsectionup
\subsection{The importance of empathy in intent inference}
\cutsubsectiondown
\begin{figure*}[ht!]
    \centering
    \includegraphics[width=1.0\textwidth]{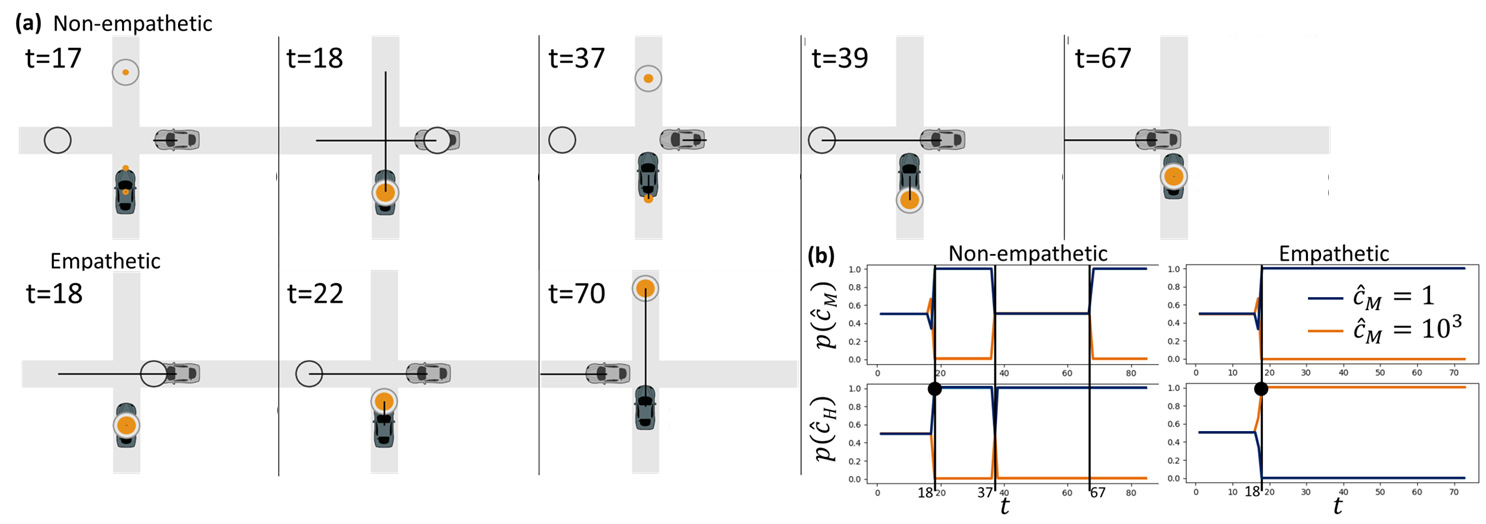}
    \vspace{-25pt}
    \caption{(a) Interactions for the empathetic and non-empathetic cases. (b) Intent inference ($\hat{c}_H$ by M and $\hat{c}_M$ by H) along time.}
    \label{fig:compare}
    \vspace{-15pt}
\end{figure*}
We now investigate the importance of empathy in intent inference using a similar experimental setting as above. We show that if agent $i$ is modeled to believe that agent $j$ already knows its true intent, then $i$ will create false estimation of $j$'s intent. \textbf{Experiment setup}: In the same intersection scenario, we use a reactive M vs. a reactive H setting with $c_M = 1$ and $c_H = 10^3$. In the non-empathetic case, we fix $\tilde{c}_M = 1$ for M's inference processes, and keep H as an empathetic agent. \textbf{Analysis}: Fig.~\ref{fig:compare} summarizes the differences between the non-empathetic and empathetic cases. In the physical space, we observe that M in the non-empathetic case iterates between a high and a low speed, similar to the reactive vs. reactive case with $c_M = c_H = 1$, until $t=37$ when M realizes that H might be more aggressive than it thought, and starts to back off. In the empathetic case, this abrupt action change does not happen. Instead, M realizes the aggressiveness of H earlier, and chooses to stop before it enters the interaction zone.

This difference in the physical space is caused by that of the intent inference, see Fig.~\ref{fig:compare}b. We will focus the discussion at $t=17$ where the two cases starts to depart from each other.
At this time step, H takes $\xi_H =1$ and M chooses to stop for both cases. The equilibrium sets for all intent combinations are summarized in Table~\ref{tab:intent}. In the empathetic case, based on Eq.~\eqref{eq:intent}, we have $p(\tilde{c}_M = 10^3,\hat{c}_H=10^3,t=18) = 1$ since this is the only combination that explains $\xi_H = 1$. This leads to $p(\hat{c}_H = 10^3,t=18) = 1$. For the non-empathetic case, since M believes that H knows about its non-aggressiveness, we have $p(\tilde{c}_M = 1,\hat{c}_H = 1,t=18) = 1$. This is because with $\tilde{c}_M = 1$, the only equilibrial motions of H are $\xi_H = 0$ and $\xi_H = 5$, between which the former is closer to the observation. This misinterpretation led to $p(\hat{c}_H = 1,t=18) = 1$, and M's more advantageous motion in the following steps.
\begin{table}[]
    \centering
    \caption{Equilibrium sets at $t=18$, reactive M and H, $c_M = 1$, $c_H=10^3$}
    \vspace{-10pt}
    \begin{tabular}{|c|c|c|c|}
        \hline
        $c_M$ & $c_H$ & $\xi_M$ & $\xi_H$ \\
        \hline
        $1$ & $1$ & $\xi_M = 0$ & $\xi_H = 0$ \\
        \hline
        $1$ & $10^3$ & $\xi_M = 0$ & $\xi_H = 5$ \\
        \hline
        $10^3$ & $1$ & $\xi_M = 5$ & $\xi_H = 0$ \\
        \hline
        $10^3$ & $10^3$ & $\xi_M = 1$ & $\xi_H = 1$ \\
        \hline
    \end{tabular}
    \label{tab:intent}
    \vspace{-20pt}
\end{table}
The above analysis showed that by fixing $\tilde{c}_M$, the agent excludes plausible explanations of others' behavior.
\section{Conclusion and Future Work}
\label{sec:discussion}
This paper investigated an empathetic intent inference algorithm for a non-collaborative interaction between two agents, and evaluated motion planning formulations that are enriched by the inference, with respect to social gracefulness and efficiency.  

We now discuss limitations of the presented study and propose future directions. 
\paragraph{The inconsistency between intent decoding and motion planning} One may notice that during motion planning, $i$ incorporates the uncertainty of the inference of $j$'s future motion. However, in intent decoding, $i$ assumes that $j$ uses a one-hot estimation of $i$'s intent ($\tilde{c}_M$) in its planning. 
To fix this inconsistency, $i$ would need to model $j$ to have considered a distribution of intents of $i$ in $j$'s planning, leading to an inference of the distribution of the distribution (of $\tilde{c}_i$) by $i$. 

\paragraph{Provable necessity and sufficiency of empathy} We demonstrated in Sec.~\ref{sec:case} a case where non-empathetic inference leads to false estimation of others' intent. However, we have not yet investigated the necessary and sufficient conditions under which empathy will be necessary for the inference to achieve correct convergence. Or mathematically, what are the necessary and sufficient conditions to have $p(\tilde{c}_i = c_i, \hat{c}_j = c_j; t) = 0$ and $p(\tilde{c}_i = c_i', \hat{c}_j = c_j; t) > 0$ for some $c_i' \neq c_i$? 

\paragraph{Inference of planning strategy} The above discussion suggests that due to the variation of planning strategies, intent inference at a parametric level may not be effective when one has a wrong guess of the others' strategy. The question is then how the inference of strategies can be incorporated, for example to differentiate a proactive agent that pretends to be aggressive and an aggressive agent. In the discrete setting as presented in this paper, the inference can be done by enumerating over all candidate strategies. This, however, will not elegantly accommodate the estimation of hyper-parameters such as $\beta$. Another potential approach is to introduce a meta-objective as a weighted sum of loss functions collected from all strategies, and to infer the strategy by estimating the weights of this meta-objective, in addition to the intent parameters.  

\paragraph{Computational challenges}
Last but not least, extending the proposed decoding and planning algorithms to continuous domains will be necessary for their scalability in multi-vehicle coordination scenarios. However, doing so will introduce computational challenges due to the non-convexity of the inference and planning problems. In addition, the incorporation of a high-dimensional distribution of inferred motions in motion planning can be intractable. We expect scenario-specific solutions to these challenges. 


\newpage
\bibliography{ref}
\bibliographystyle{IEEEtranS}
\end{document}